%% file: main.tex
\newif\ifRAL
\RALfalse   % ICRA paper

\newif\ifTR
\TRfalse 	% Tech report not in

\newif\ifPrePrint
\PrePrintfalse

\ifRAL
  \documentclass[letterpaper, 10pt, journal, twoside]{IEEEtran}
\else
  \documentclass[letterpaper, 10pt, conference]{ieeeconf}
\fi

\usepackage[normalem]{ulem}

\makeatletter

\let\proof\@undefined
\let\endproof\@undefined
\makeatother

\usepackage{graphicx}
\usepackage{epstopdf}
\usepackage{lipsum}
\usepackage{amsthm}
\usepackage{amsfonts}
\usepackage{mathtools}  % used for \DeclarePairedDelimiter
\usepackage{bm}  % used for bold and prettier greek letters with \bm
\usepackage[dvipsnames]{xcolor}  % used for colored text

\usepackage{paralist}  % used for inparaenum env
\usepackage{tikz}
\usepackage[textsize=tiny]{todonotes}
\usepackage{caption}  % used to set caption size
\usepackage{subcaption}  % used for subfigures
\usepackage{cite}
\usepackage{booktabs}  % better table hlines
\usepackage{siunitx}  % for \num
\usepackage[hang,flushmargin]{footmisc}  % footnote indentation, hang flushes the marker, flushmargin flushes the text
\usepackage[bookmarks]{hyperref}  % used to get pdf with links and toc
\hypersetup{
    colorlinks,
    citecolor=black,
    filecolor=black,
    linkcolor=black,
    urlcolor=black,
    pdfauthor={},
    pdfsubject={},
    pdftitle={}
}
\usepackage{ulem}
\usepackage{xspace}  % used for \xspace in macros (smart spacing)
\IEEEoverridecommandlockouts  % used for \thanks

\graphicspath{{imgs/}}
\def\baselinestretch{0.975}

\title{\LARGE \bf \acs{NMPC} for Deep Neural Network-Based Collision Avoidance exploiting Depth Images}
\author{Martin Jacquet$^1$, Kostas Alexis$^1$
	\thanks{$^1$Autonomous Robots Lab, Norwegian University of Science and Technology (NTNU), Trondheim, Norway,
    {\tt \footnotesize
        \href{mailto:martin.jacquet@laas.fr}{martin.jacquet@ntnu.no},
    	\href{mailto:antonio.franchi@laas.fr}{konstantinos.alexis@ntnu.no}}
    }
    \thanks{This work was partially supported by the European Commission Horizon project DIGIFOREST (EC 101070405).}
    \ifRAL
        \thanks{Manuscript received: September 9, 2021; Revised: December 10, 2021, Accepted: January 5, 2022.}
        \thanks{This paper was recommended for publication by Editor Pauline Pounds upon evaluation of the Associate Editor and Reviewers' comments.}
        \thanks{Digital Object Identifier (DOI): 10.1109/LRA.2022.3143218}
    \else
    	{} % nothing
    \fi
}

\ifPrePrint
	\makeatletter
    \ifRAL
        \def\ps@IEEEtitlepagestyle{
    		\def\@oddfoot{}
            \def\@evenfoot{}
    		\def\@oddhead{\textcolor{red}{\sf Preprint version\hfill}}
    		\def\@evenhead{}
    	}
    \else
        \def\ps@tuall{
            \def\@oddfoot{}
            \def\@evenfoot{}
            \def\@oddhead{\textcolor{red}{\sf Preprint version\hfill}}
            \def\@evenhead{}
        }
    \fi
	\def\ps@headings{
		\def\@oddfoot{\textcolor{red}{\sf Preprint version\hfill}}
        \def\@evenfoot{\textcolor{red}{\sf Preprint version\hfill}}
		\def\@oddhead{}
        \def\@evenhead{}
	}
	\makeatother
\fi

\input{0_abbrev}

\input{0_glossary}

\begin{document}
\maketitle

%%% ABSTRACT %%%%%%%%%%%%%%%%%%%%%%%%%%%%%%%%%%%%%%%%%%%%%%%%%%%
\begin{abstract}
    This paper introduces a \nmpc framework exploiting a Deep \acl{NN} for processing onboard-captured depth images for collision avoidance in trajectory-tracking tasks with \acsp{UAV}.
    The network is trained on simulated depth images to output a collision score for queried 3D points within the sensor field of view.
    Then, this network is translated into an algebraic symbolic equation and included in the \nmpc, explicitly constraining predicted positions to be collision-free throughout the receding horizon.
    The \nmpc achieves real time control of a \acs{UAV} with a control frequency of 100Hz.
    The proposed framework is validated through statistical analysis of the collision classifier network,
    as well as Gazebo simulations and real experiments to assess the resulting capabilities of the \nmpc to effectively avoid collisions in cluttered environments.
    The associated code is released open-source.
\end{abstract}

%%% KEYWORDS %%%%%%%%%%%%%%%%%%%%%%%%%%%%%%%%%%%%%%%%%%%%%%%%%%%
\ifRAL
	% Decomment only if accepted to RAL
\else
	{} % nothing
\fi

%%% INTRODUCTION %%%%%%%%%%%%%%%%%%%%%%%%%%%%%%%%%%%%%%%%%%%%%%%
\acresetall  % reset acronyms for mainmatter
\section{Introduction}\label{sec:intro}
% Decomment àonly if accepted to RAL
% \ifRAL
%     \IEEEPARstart{U}{ncrewed }
% \else
%     {Uncrewed }
% \fi
\ars are increasingly used in a large range of autonomous applications, from aerial monitoring or exploration to working in high-risk places or human-denied areas,
or in tasks such as
search-and-rescue~\cite{Tian20},
subterranean exploration~\cite{Dang20}
and indoor building inspection~\cite{Petracek20}.
Furthermore, the increasing efficiency and decreasing weight of the available sensors and computation units allowed the deployment of recent efficient computer vision algorithms on UAVs~\cite{Zhang19, Akbari21}.
However, full autonomy of \ars in unknown and possibly cluttered environments remains a challenging task, as localization and mapping -- relying only on onboard sensors -- is subject to significant noise and drift, especially in visually degraded environments~\cite{Khattak20}.
Although there exist active sensors, such as \acsp{LIDAR}, allowing precise high-density mapping which can be exploited by a collision-avoidance planner,
those are low-frequency and heavy sensors that are not suitable for fast flights.
Furthermore, the high computational requirements of dense-map planning algorithms are subsequently limiting the velocity of \ars.
This issue becomes prominent in time-critical applications,
and greatly reduces the distance coverage for a given battery time, \eg in exploration tasks.

Thus, recent works~\cite{Lopez17, Loquercio18, Loquercio21, Tolani21, Kahn21, Ugurlu22, Hoeller21, Nguyen22} are taking a different approach by relying purely on sensor data and local estimates for short-term collision avoidance, tackled at the level of the controller.
The main challenge in such methods lies in the high dimension of sensor data, rendering classical approaches (\eg,~\cite{Lopez17}) difficult to implement.
Instead, some of the aforementioned works have proven that \nns are efficient tools to deal with such large input spaces.
In this context, the favored sensors are depth images, which are both easy to embed onboard and easier to simulate than classical RGB cameras.
Using such data allows to train on large batches of simulated images with a relatively small sim-to-real gap.
Some recent works are also investigating learning navigation from RGB images~\cite{Tolani21, Kahn21}.

Common approaches to such sensor-based navigation policies are \acl{IL} and \ac{RL}.
The former makes use of a privileged policy~\cite{Loquercio18, Loquercio21, Tolani21} (\eg exploiting map knowledge) that is imitated by a \nn-based controller accessing only sensor measurements.
Such approaches are however limited by the availability of such privileged policies,
and do not generalize well to unknown situations.
\ac{RL} methods~\cite{Kahn21, Ugurlu22} on the other hand are attracting more and more attention with the surge of efficient simulation environments,
and recent demonstration of performances, \eg in drone racing~\cite{Kaufmann23}.
\ac{RL} provides efficient end-to-end control schemes and is successfully employed for sensor-based collision-free navigation~\cite{Kahn21, Hoeller21, Ugurlu22}.
Recent evidence~\cite{Tan23} show that under some assumptions on the reward function,
such methods also provides tools for safety certification, through \ac{CBF},
paving the path for further research on safety-oriented \ac{RL}.

In the scope of collision avoidance,
\nns have also been used in combination with classical planning or control methods.
Contrary to end-to-end methods, it enables a more transparent correspondence with the modular approach which has largely governed robotics research over the past years.
Corollary, it allows more control on the framework by allowing evaluation the performances of the individual blocks.
In~\cite{Dawson22}, a \nn exploits \acs{LIDAR} data to synthesize a \ac{CBF} that certifies safety of the system in the current observable environment,
from which safe commands are derived using classical control.
In~\cite{Nguyen22}, a \nn is used to predict collision scores of some motion primitives based on partial current state estimates,
in a receding horizon fashion.
The predictive aspect is thus handled by the \nn which approximates collision rollouts.
However, this method is limited by the choice of sampling of trajectories,
while predictive optimal local planners or controllers, such as \nmpc, provide more flexibility.

However, \nmpc has not been used in combination with \nns for sensor-based navigation,
but mainly for learning the dynamics residual to reduce the imprecision of the models.
To this end, recent paradigms have been proposed~\cite{Williams17, Chee22, Salzmann23} to integrate \nn in \nmpc schemes.
A first approach is to leverage \nns to guide sampling-based \mpc~\cite{Williams17}.
On the other hand, in \cite{Salzmann23}, the authors propose a so-called Neural \mpc, a \mpc using deep learning models in its prediction step.
The \nn is directly embedded into the optimal problem as an algebraic symbolic equation, enabling gradient-based optimization.

In this work, we propose to combine the local planning aspects of \nmpc schemes with deep \nn to process input data,
achieving sensor-based collision avoidance in real time.
A deep \nn is designed for collision prediction using depth images.
This \nn is queried for position states and outputs a collision score.
Using the same approach as \cite{Salzmann23}, this \nn is integrated into the \nlp equations,
constraining predicted positions to remain collision-free.
The structure of the \nn is designed such that the size of the matrices defining the symbolic neural prediction constraint is maintained small,
enabling fast optimization, and consequently real time control.

The paper is organized as follows:
first, the modeling of the \ar is formally introduced.
Then, the deep \nn architecture is presented in \sect{sec:nn} and the training methodology is defined,
leading to the definition of the collision-aware \nmpc in \sect{sec:mpc}.
Finally, the method is evaluated both in simulations and real experiments in \sect{sec:valid},
before concluding.

%%% SEC %%%%%%%%%%%%%%%%%%%%%%%%%%%%%%%%%%%%%%%%%%%%%%%%%%%%%%%%
\section{Modeling}\label{sec:model}
We define the world inertial frame $\frameV{W}$, with its origin $O\us{W}$ and its axes $\vect{x}\us{W}, \vect{y}\us{W}, \vect{z}\us{W}$.
Following the same convention, the body frame of a \ar and the depth camera frame are respectively denoted $\frameV{B}$ and $\frameV{C}$.

The \ar is defined as a rigid body %of mass $m$,
centered in $O\us{B}$, and actuated by typically $4$ or $6$ co-planar propellers.
Its position \wrt the $\frameV{W}$ is denoted by $\ts{W}\vect{p}\us{B}$
and the rotation matrix from $\frameV{B}$ to $\frameV{W}$ is denoted by $\ts{W}\mat{R}\us{B}$; and similarly for all the other frame pairs.
The unit quaternion representation of the rotation $\ts{W}\mat{R}\us{B}$ is denoted $\ts{W}\vect{q}\us{B}$.

The \ar is assumed to embed a front-facing depth camera,
rigidly attached such that $\ts{B}\vect{p}\us{C}$ and $\ts{B}\mat{R}\us{C}$ are constant and known.
Its \fov, denoted $\mathbb{F}$, is a pyramidal volume described by two angles $\alpha\us{V}$ and $\alpha\us{H}$ and a height $d_{\text{max}}$, respectively describing the halved vertical and horizontal angular apertures and maximum sensing depth.
Its principal axis is aligned with $\vect{z}\us{C}$.
This camera provides, at a given frequency $f\us{C}$,
a depth image $I(t)$, rasterized in $n\us{V}\times n\us{H}$ pixels, whose scalar values the depth of the closest objects in the corresponding angular sector of $\mathbb{F}$.
Pixel depth values are normalized by $d_{\text{max}}$ such that they range in $\intervei{0}{1}$.

The system is expected to handle obstacles through the images captured by the depth camera.
Therefore, the motion of the \ar must occur within the camera \fov to enable obstacle avoidance,
implying that only forward motion with pitching and yawing are allowed.
The dynamics of the \ar are accordingly restricted to those of a non-holonomic system.

The system state is described by the vector
\begin{equation}
    \vect{x} = [\vect{p}\transp ~ \vect{q}\transp ~ v_x]\transp \in\R^{8},
\label{eq:model:x}
\end{equation}
where $\vect{p} = \ts{W}\vect{p}\us{B}$, $\vect{q} = \ts{W}\vect{q}\us{B}$,
$v_x$ is the forward velocity of $O\us{B}$, expressed in $\frameV{B}$.
The system input variables are
\begin{equation}
    \vect{u} = [a_x ~ \omega_y ~ \omega_z]\transp \in\R^{3},
\label{eq:model:u}
\end{equation}
where $\omega\us{y}$ and $\omega\us{z}$ are pitching and yawing rates of $\frameV{B}$ \wrt $\frameV{W}$, expressed in $\frameV{B}$,
and $a_x$ is the forward acceleration of of $O\us{B}$, also expressed in $\frameV{B}$.
% It is assumed that lower and upper bounds on $\gamma$ and $\vecg{\omega}$, are known, or obtained through an identification campaign.
% Although this choice of intermediate input may limit the capability of the model to capture the full actuation span of the \ar compared to motor-level \nmpc~\cite{Bicego20},
% it is a reasonable description of its motion when dealing with non-agile maneuvers, as it is the case during, \eg, exploration tasks.

Accordingly, the system kinematics and dynamics are defined by
\begin{subequations}
    \begin{align}
        \dot{\vect{p}} &= \ts{W}\mat{R}\us{B}\bmat{0 ~ 0 ~ v_x}\transp, \\
        \dot{\vect{q}} &= \frac{1}{2}~\vect{q}\otimes\bmat{0 ~ 0 ~ \omega_y ~ \omega_z}\transp, \\
        \dot{v_x} &= a_x,
    \end{align}
    \label{eq:model:dyn}%
\end{subequations}
where $\otimes$ denotes the Hamilton product of two quaternions.

%%% SEC %%%%%%%%%%%%%%%%%%%%%%%%%%%%%%%%%%%%%%%%%%%%%%%%%%%%%%%%
\section{Deep Neural Collision Predictor}\label{sec:nn}
Given a depth image $I$ and a 3D point $\vect{a}$, there exists a mapping
\begin{equation}
    \begin{split}
        g~\colon ~\mathbb{I}\times\mathbb{F} ~ & \rightarrow \intervint{0}{1}\\
        ~~(I, \vect{a}) ~ & \mapsto ~~c
    \end{split}
\label{eq:nn:col}
\end{equation}
where $\mathbb{I} = \intervei{0}{1}^{(n\us{V}\times n\us{H})}$ is the depth image space,
and $c$ is a Boolean value describing whether $\vect{a}$ is in collision with an obstacle in $I$.
Note that points that fall behind obstacles are not visible from the depth camera, and thus are considered to be in collision.
Intuitively, this implies that any point that is not directly visible potentially collides with an obstacle, and therefore is conservatively classified as being in collision.
This assumption is also used to extend the domain of definition of $g$ to $\mathbb{I}\times\R^3$ with $g(I, \vect{a}) = 1, ~\forall \vect{a}\notin\mathbb{F}$.

Such function $g$ is discontinuous and challenging to write in closed form,
therefore it does not allow any gradient-based optimization in order to let a \nmpc avoid collisions throughout its receding horizon.

Instead, a continuous parametric approximation of $g$ can be defined as
\begin{equation}
    \begin{split}
        g_{\vecg{\theta}}~\colon ~\mathbb{I}\times\R^3 ~ & \rightarrow \interv{0}{1}\\
        ~(I, \vect{a}) ~~ & \mapsto ~~\hat{c}
    \end{split}
\label{eq:nn:out_of_fov}
\end{equation}
where $\vecg{\theta}$ is a set of parameters and $\hat{c}$ is an approximate collision score such that
\begin{equation}
    \forall (I,\vect{a}),~~
    \begin{cases}
        \hat{c} \approx 1 ~~\text{if}~ c = 1, \\
        \hat{c} \approx 0 ~~\text{if}~ c = 0.
    \end{cases}
\label{eq:nn:col_approx}
\end{equation}

We note that the method itself is not restricted to using depth images.
Other depth-based sensors such as \acp{LIDAR} could be employed similarly.

%---------------------------------------------------------------
\subsection{\acl{NN} Architecture}\label{subsec:nn:arch}
Our objective is to design a deep \nn that learns an accurate approximation $g_{\vecg{\theta}}$ of $g$.

\begin{figure}[t]
\centering
    \scalebox{1.05}{
        \def\svgwidth{\columnwidth}\footnotesize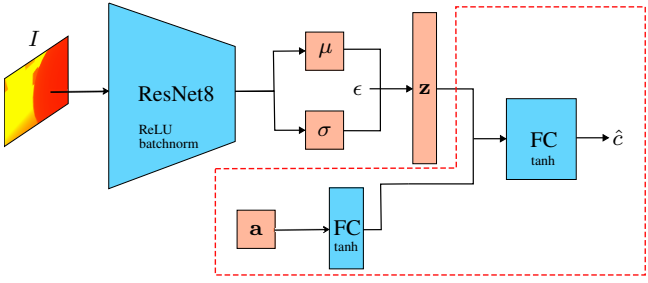
    }
    \caption{Architecture of the classifier network.
    The image is encoded into a latent representation $\vect{z}$,
    which is sampled from a learned distribution $\vect{z}\sim\mathcal{N}(\mu,\sigma)$,
    using a randomness variable $\epsilon$.
    The \ac{FC} network is a coordinates-based \nn, which outputs a predicted collision score $\hat{c}$ for input points $\vect{a}\in\R^3$ in the volumetric map described by $I$.
    The red lines highlight the part of the network to be embedded into the \nmpc.}
    \label{fig:nn_arch}
\end{figure}

This network is designed after Variational Encoder-Decoder architectures,
depicted in~\fig{fig:nn_arch}.
The variational encoder is a \cnn that processes the input image $I$ into a latent representation $\vect{z}\sim\mathcal{N}(\mu,\sigma)$ of reduced dimensionality $n\us{Z}$.
This latent vector is concatenated with the $3$D point $\vect{a}$
and processed by a \ac{FC} network which outputs the desired scalar approximate collision score $\hat{c}$.

Similar to~\cite{Loquercio18, Nguyen22}, we chose a ResNet-8 architecture for the \ac{CNN}, using ReLU activations,
batch normalizations and dropouts.
The output volume of the last layer is passed through an average pooling layer,
before flattening and dimension reduction with a pair of \ac{FC} layer to compute the mean $\mu$ and \std $\sigma$ of the latent encoding.

The \ac{FC} network is trained as a coordinate-based \ac{FC} network~\cite{Mescheder19, Tancik20},
that takes as input a 3D point $\vect{a}$ and $\vect{z}$, and outputs $\hat{c}$.
This network is chosen to be relatively shallow ($3$ hidden layers) to reduce gradient decay issues,
and the layer widths are maintained small ($\le128$) for reducing the size of $\vecg{\theta}$ and consequently the solving time of the \nmpc.
It uses $\tanh$ activations, as the resulting \ac{FC} network needs to be fully differentiable \wrt $\vect{a}$ to allow gradient-based optimization.
A final sigmoid unit is employed to constrain its output in $\interv{0}{1}$.

In order to reduce the dimensionality bias between $\vect{a}$ and $\vect{z}$ as input of the \ac{FC} network (typically, $3$ against $128$),
we process $\vect{a}$ with a first \ac{FC} layer before the concatenation with the latent vector.

%---------------------------------------------------------------
\subsection{Training Methodology}\label{subsec:nn:train}
The overall network is trained as a classifier,
using a \ac{BCE} loss $\mathcal{L}_\text{BCE}$ between the network output $\hat{c}$ and the actual label $c$, which is computed using an algorithmic implementation of \eqn{eq:nn:col}.
It is weighted such that collision samples are given more importance in the training,
in order to reduce the false negative rate.
This renders the resulting classifier more conservative but minimizes the likelihood of unpredicted collisions.
A $\beta$-scaled Kullback-Leibler divergence metric $\mathcal{L}_\text{KL}$~\cite{Higgins17} is used to enforce that $\vect{z}$ follows a proper normal distribution fitting the true posterior,
as commonly done for variational encoders.

The training loss function $\mathcal{L} = \mathcal{L}_\text{BCE} + \mathcal{L}_\text{KL}$ is given by
\begin{align}
        \mathcal{L}_\text{BCE} &= \frac{-1}{nb}\sum_{i=1}^b{\sum_{j=1}^n{\lambda_1c_{ij}\log\hat{c}_{ij}+\lambda_0(1-c_{ij})(1-\log\hat{c}_{ij})}}, \\
        \mathcal{L}_\text{KL}  &= -\frac{\beta_{\text{norm}}}{2}\sum_{i=0}^b{1+\log(\sigma_i^2)-\mu_i^2-\sigma_i^2},
\end{align}
where $b$ is the number of batch elements,
$n$ is the number of points sampled per image,
subscripts $i$ and $j$ denotes quantities being computed from inputs $I_j$ and $\vect{a}_j$,
$\lambda_0$ and $\lambda_1$ are respectively weights for the $0$ and $1$ classes,
and $\beta_{\text{norm}}$ is a scaling factor computed from $n\us{Z}$, $n\us{H}$ and $n\us{V}$, according to~\cite{Higgins17}.

The \nn is trained entirely on simulated depth images, obtained using Aerial Gym~\cite{Kulkarni23b}.
The generated environments are sampled to be slightly cluttered,
with 20 objects of medium sizes randomly placed (and oriented) in $5\unit{m^3}$,
along with $2$ long pillars, also randomly placed, and $4$ walls.
A random $6$D pose of the camera is sampled within each simulated environment.
Training dataset totalizes $250$k images before augmentation (randomized flipping, shifting, and noising).

To account for the fact that the network infers the collision label for a single 3D point instead of the full \ar volume,
the training labels are computed on depth images processed such that obstacles are ``inflated'' by the radius of the drone~\cite{Kulkarni23c},
adding a constant unknown bias to be learned.

% The \nn is trained such that the latent representation of an image is the same for all the points sampled in this image within a batch.
The input positions are sampled uniformly in spherical coordinates,
in a volume that is larger than the \fov $\mathbb{F}$,
to ensure control on the \nn output at the boundary, as per \eqn{eq:nn:out_of_fov}.
In~\cite{Mescheder19}, the authors showed that uniform sampling provides the best consistency in a similar $3$D volume reconstruction task,
as other sampling methods tends to introduce a bias in the model.
We chose a high number of points sampled per image (\eg, $10^4$),
in order to statistically ensure that some sampled points fall into small objects during training.
We note that the actual input $\vect{a}$ of the \nn is scaled by $d_{\text{max}}$, $\alpha\us{H}$ and $\alpha\us{V}$, such that
\begin{equation}
    \vect{a} = \frac{1}{d_{\text{max}}}\bmat{
        \frac{x}{\tan\alpha\us{H}},~
        \frac{y}{\tan\alpha\us{V}},~
        z
    }\transp.
\label{eq:nn:scale_inputs}
\end{equation}

Since the collision classifier is to be included as a safety constraint in the \nmpc,
it is mandatory that the current position of the drone is classified as free, \ie that there always exists a safe solution for the \ar.
Moreover, for numerical stability when hovering, the positions close to the current one must also be classified as safe.
Therefore, a ball of small radius (typically, a couple of centimeters) is defined around $O\us{C}$ within which all states are labeled as safe.
Throughout training, points are sampled in this ball for each image,
ensuring that the \nn is well conditioned in this volume.

%%% SEC %%%%%%%%%%%%%%%%%%%%%%%%%%%%%%%%%%%%%%%%%%%%%%%%%%%%%%%%
\section{\nmpc with Deep Collision Prediction}\label{sec:mpc}
%---------------------------------------------------------------
\subsection{Neural \nmpc}\label{subsec:mpc:def}
In order to track position or velocity trajectories without colliding with surrounding obstacles,
the collision prediction \nn is integrated into the \nmpc scheme,
as constraints on the \ar position.
The prediction step of the \nmpc is independent of the \ac{CNN} part of the network,
as the image input $I$ is fixed for a given optimization loop.

Therefore, we must write $\hat{c}$ as a closed-form (parametric) function of the \nmpc state vector $\vect{x}$.
The $3$D input of the \nn for prediction steps of the \nmpc is the position of the camera at a given time $t$, $O_{C}(t)$,
expressed in the frame $\frameV{C_0} = \frameV{C}(t_0)$ at which the depth image was captured.
It is a function of $\vect{x}$ given by:
\begin{equation}
\begin{split}
    \ts{C_0}\vect{p}\us{C} &=
        \ts{C_0}\mat{R}\us{W}\ts{W}\vect{p}\us{C} + \ts{C_0}\vect{p}\us{W} \\
    ~&=
        \ts{W}\mat{R}\us{C_0}\transp
        (\ts{W}\mat{R}\us{B} \ts{B}\vect{p}\us{C} + \ts{W}\vect{p}\us{B} - \ts{W}\vect{p}\us{C_0})
\end{split}
\end{equation}
where $\ts{B}\vect{p}\us{C}$ is fixed and known,
$\ts{W}\vect{p}\us{B}$ and $\ts{W}\mat{R}\us{B}$ are functions of $\vect{x}$,
and $\ts{W}\vect{p}\us{C_0}$ and $\ts{W}\mat{R}\us{C_0}$ are parameters computed from the pose of the \ar at the moment the depth image is captured.

Then, for a given depth image $I$ captured at $t=t_0$, we denote $g_{\vecg{\theta},\vect{z}}(\ts{C_0}\vect{p}\us{C})$ the quantity to constrain,
that is, the output of the \ac{FC} part of the \nn evaluated on the latent representation $\vect{z}$ of $I$, and the $3$D position of the camera over the receding horizon, expressed in $\frameV{C_0}$.

When evaluating the collision avoidance constraint,
in order to make it convex and avoid vanishing gradients,
the terminal sigmoid activation is replaced by an exponential unit,
utilizing that
\begin{equation}
    \forall x \in\R, ~~\text{sigmoid}(x) < 0.5 ~\Leftrightarrow~ \frac{e^x}{2} < 0.5.
\end{equation}

We remark that even though position information is required for collision avoidance,
contrary to partial-state-based navigation policies~\cite{Nguyen22},
the prediction is performed locally, \wrt $\frameV{C_0}$ which moves with the \ar at high frequency (typically, $60$\unit{Hz}),
therefore alleviating the drift issues inherent to map-based collision avoidance.

%---------------------------------------------------------------
\subsection{\acl{NLP}}\label{subsec:mpc:nlp}
The collision-aware trajectory-tracking \nmpc objective is defined by the minimization of the weighted square norm of an output vector $\vect{y}(\vect{x})$ \wrt its reference $\vect{y}_r$,
denoted $\norm{\vect{y}({\vect{x}})-\vect{y}_{r,k}}^2_\mat{W}$.
Such reference trajectory $\vect{y}_r$ is typically defined as sequences of position waypoints or velocity references, depending on the considered task.
It is provided by a higher-level planner, \eg based on a goal position to reach.

Additionally, the minimization of $g_{\vecg{\theta},\vect{z}}(\ts{C_0}\vect{p}\us{C})$ is included to the \nmpc cost function,
in order to guide the \nmpc toward a solution that satisfies the obstacle avoidance constraint.

The discrete-time \ac{NLP} over the receding horizon $T$,
sampled in $N$ shooting points, at a given instant $t$,
given a depth image captured at $t_0 \le t$ and compressed into a latent vector $\vect{z}$,
is expressed as
\begin{subequations}
\begin{equation}
    \min_{\substack{\vect{x}_0\dots\vect{x}_N \\ \vect{u}_0\dots\vect{u}_{N-1}}}
        ~\sum_{k=0}^{N}{\norm{\vect{y}({\vect{x}_k})-\vect{y}_{r,k}}^2_\mat{W}
        + w~g_{\vecg{\theta},\vect{z}}(\ts{C_0}\vect{p}\us{C,k})
        }\label{eq:mpc:nlp:cost_function}%
\end{equation}%
\begin{align}
    s.t. ~~~&\vect{x}_0 = \vect{x}(t) \\
    &\vect{x}_{k+1} = \mathbf f(\vect{x}_k,\vect{u}_k), &{\scriptstyle k\in\{0, N-1\}}\\
    &0 \le v_{x,k} \le \overline{v_x}, &{\scriptstyle k\in\{0, N\}}\label{eq:mpc:nlp:x_constr}\\
    &\underline{\vect{u}} \le \vect{u}_k \le \overline{\vect{u}}, &{\scriptstyle k\in\{0, N-1\}}\label{eq:mpc:nlp:u_constr}\\
    &g_{\vecg{\theta},\vect{z}}(\ts{C_0}\vect{p}\us{C,k}) \le 0.5, &{\scriptstyle k\in\{0, N\}}\label{eq:mpc:nlp:avoidance}
\end{align}
\label{eq:mpc:nlp}%
\end{subequations}

where $\vect{x}(t)$ is the state estimate at $t$,
$\vect{f}$ synthetically denotes the dynamics defined in \eqn{eq:model:x},
$\mat{W}$ and $w$ are tunable weights,
and $\underline{\bullet}$ and $\overline{\bullet}$ denotes mission-related lower and upper bounds on the \nmpc state and inputs.

%%% SEC %%%%%%%%%%%%%%%%%%%%%%%%%%%%%%%%%%%%%%%%%%%%%%%%%%%%%%%%
\section{Validation}\label{sec:valid}
%---------------------------------------------------------------
\subsection{Collision Classifier}\label{subsec:valid:nn}
This section presents a quantitative analysis of the collision classifier.
A testing set of $5000$ simulated images are gathered,
within which $4$M points are randomly sampled ($\approx0.2$ points/cm$^3$).
Accuracy, precision, and recall of the \nn classifier are computed for each image and reported in~\tab{tab:classif_metrics}.
The threshold for computing the metrics is $0.5$,
to fit the chosen value defined as the upper bound of the \nmpc constraint.

To assess the sim-to-real gap of the method,
the \nn is also evaluated on a set of real images captured with a Realsense d455 camera.
We make use of evaluation images used in~\cite{Kulkarni23a},
\ie a dataset of 1498 images captured in confined spaces, indoor rooms,
long corridors, and outdoor environments with trees.
The systematic errors in depth images (stereo shadow) are compensated with a filling algorithm~\cite{Ku18}.

\begin{figure}[t]
\centering%
    \includegraphics[width=0.46\columnwidth]{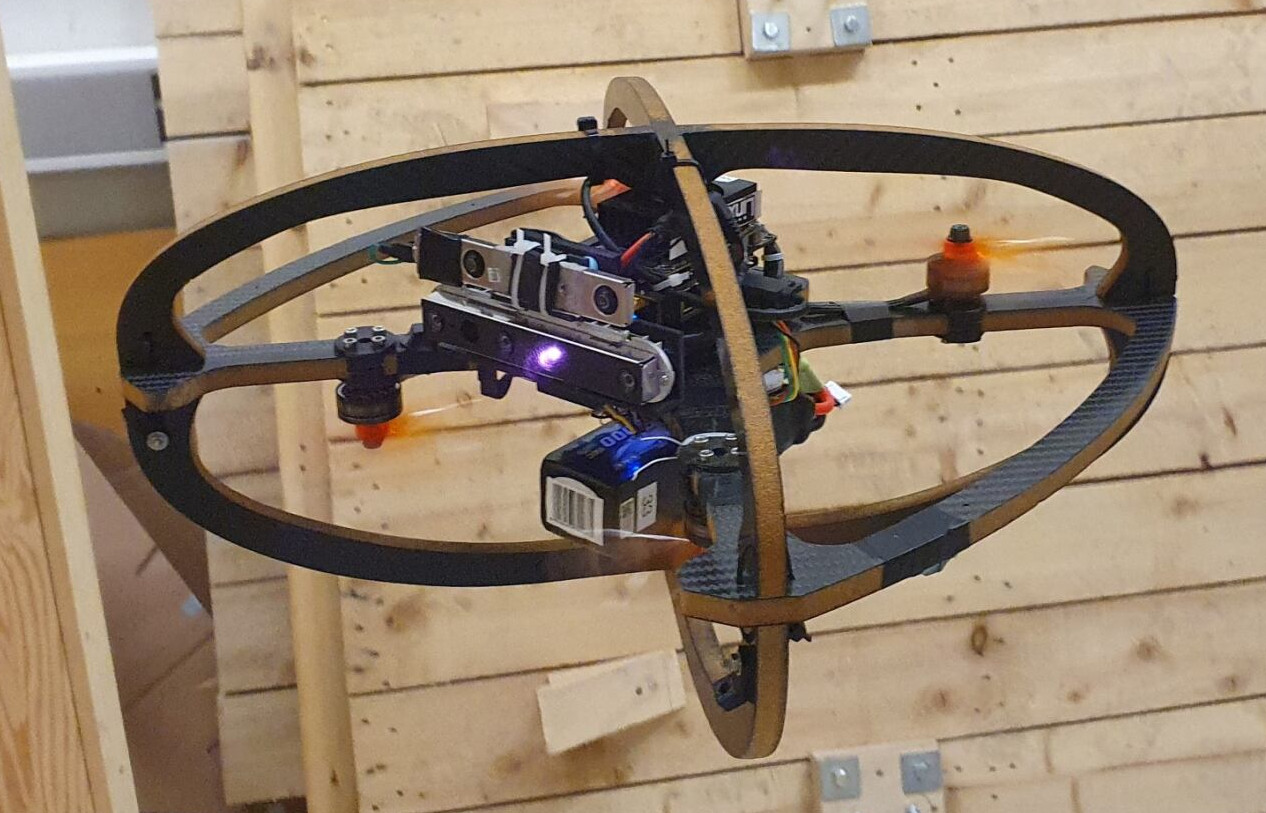}
    \hfill
    \includegraphics[width=0.525\columnwidth]{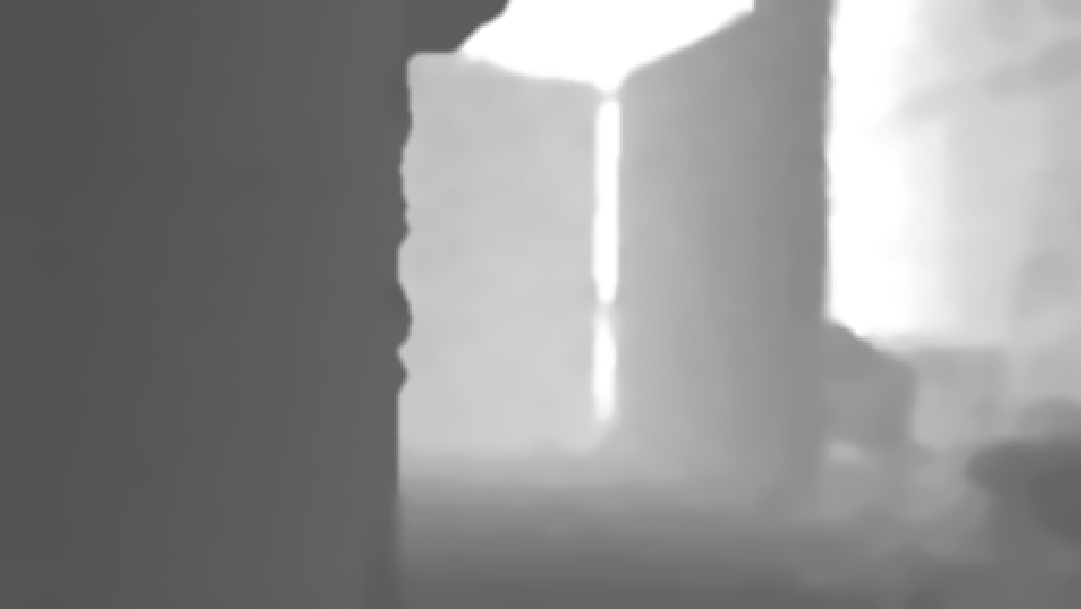}
    \caption{The \ar used in the reported experiment, LMF, and a view of the (filled) depth image from its onboard front-facing camera.}
\label{fig:lmf_view}%
\end{figure}

\begin{table}[t]
    \centering
    \resizebox{\columnwidth}{!}{
        \begin{tabular}{|c|c|c|c|c|c|c|}
        \hline
                            & \multicolumn{2}{c|}{Accuracy}   & \multicolumn{2}{c|}{Precision}    & \multicolumn{2}{c|}{Recall}       \\
        \cline{2-7}
                            & \textit{mean} & \textit{std}    & \textit{mean} & \textit{std}      & \textit{mean} & \textit{std}      \\
        \hline
        Simulated images    & 93.1\% & 4.2\%  & 86.0\% & 10.0\%  & 98.6\% & 2.9\% \\
        \hline
        Real images         & 95.4\% & 3.4\%  & 92.1\% & 6.8\%  & 98.3\% & 2.4\% \\
        \hline
        \end{tabular}
    }
    \caption{Metrics of the \nn collision classifier (mean and \std).}
    \label{tab:classif_metrics}
\end{table}

The weighting of the \ac{BCE} mentioned in~\sect{subsec:nn:train} is, as expected,
inducing a high recall of the classifier, to the detriment of precision.
The metrics computed on real images are higher than for simulated data.
This is explained by the fact that metrics are computed against the filled depth image,
which is blurred during preprocessing.
The resulting collision volume is thus smoothed and easier to approximate by the \nn.
Moreover, the real environments are inherently more structured than the chaotic synthetic randomly sampled simulation environments.
The good performances on real images demonstrate the pertinence of the simulation-trained method and its applicability to real-world scenarios,
as shown in~\sect{subsec:valid:xp}.

%---------------------------------------------------------------
\subsection{Setup}\label{subsec:valid:nn}
The proposed \nmpc is implemented in Python using Acados~\cite{AcadosLib} and Casadi~\cite{CasADiLib}.
The neural network implementation is written with PyTorch,
and it is interfaced with the \nmpc using ML-Casadi~\cite{Salzmann23}.
The \nlp is transformed into a SQP solved with a \ac{RTI} scheme.
The \nmpc inputs $\vect{u}$ is transformed into velocity commands and sent to the simulated or real system,
\eg through ROS or GenoM~\cite{Mallet10}, which handles state estimation and low-level control.
A goal waypoint is provided to the controller,
which computes before each iteration a reference velocity vector of constant norm to go toward this point.
Both in simulation and experiments, the receding horizon is set to $T=4$\unit{s}, sampled in $N=10$ points.
The code for the \nn training and inference, the \nmpc controller, and the ROS interface are released as open-source
\footnote{\tt\url{https://github.com/ntnu-arl/colpred_nmpc}}.

%---------------------------------------------------------------
\subsection{Gazebo Simulations}\label{subsec:valid:sim}
\begin{figure}[t]
\centering
    \includegraphics[width=0.99\columnwidth]{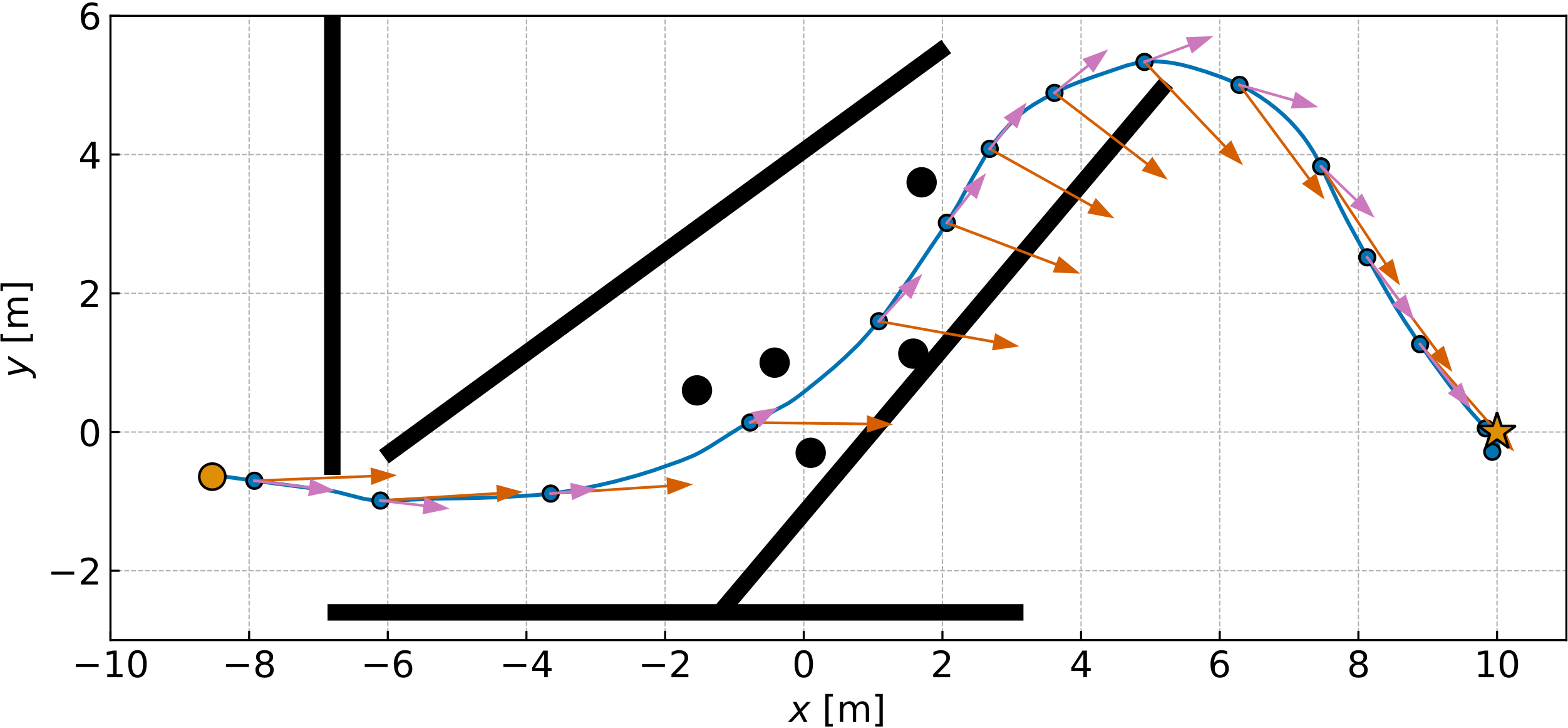}
    \caption{$(x,y)$ motion of the \ar, in blue, among the black obstacles.
    The orange circle and stars are the initial and goal positions.
    The blue dots are the position of the \ar every 2\unit{s},
    while the red and pink arrows are resp. the reference and actual velocity vector at the corresponding time.}
    \label{fig:2D_sim}
\end{figure}

\begin{figure}[t]
\centering
    \includegraphics[width=0.99\columnwidth]{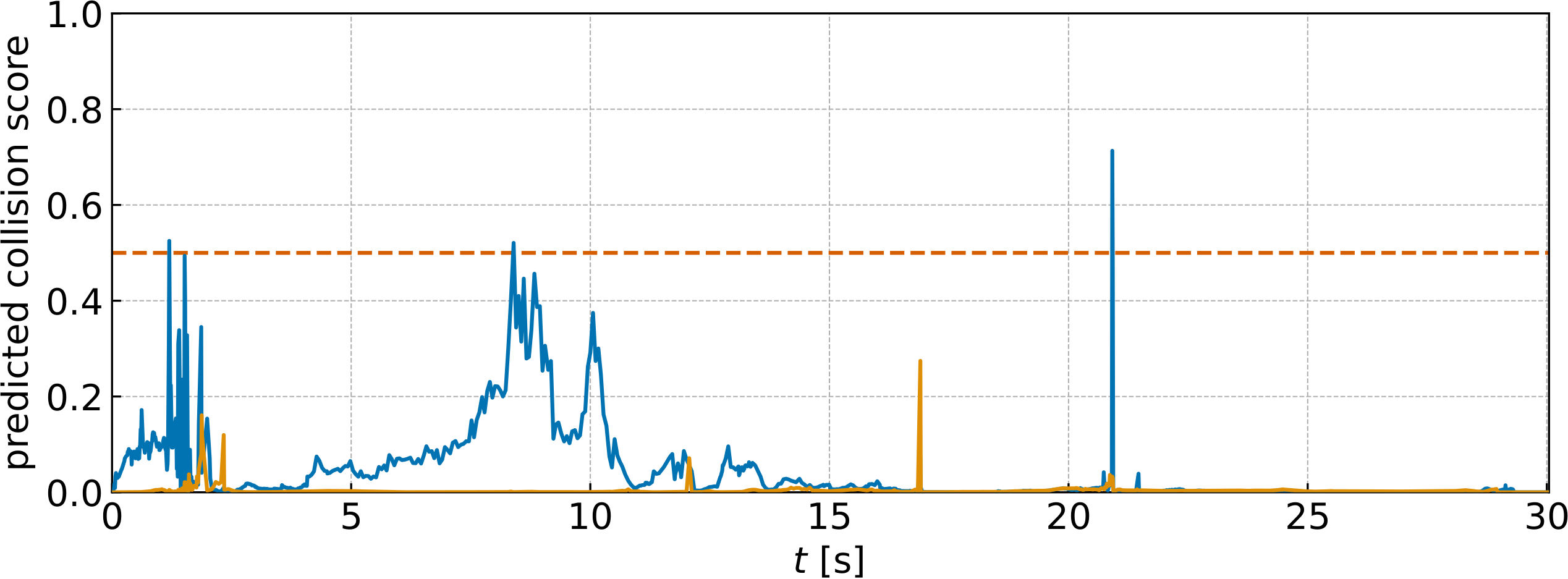}
    \caption{Predicted collision score for the first (orange) and last (blue) shooting points of the receding horizon, throughout the trajectory, with its corresponding upper bound (red).}
    \label{fig:colpred}
\end{figure}

We first present some simulation result of the proposed framework.
The \ar is tasked to reach a waypoint behind a corridor filled with pillars.
To showcase the avoidance behavior yield by the method,
the \nmpc is weighted to maintain a constant $z$, while tracking a $(x,y)$ velocity vector (of constant norm) toward the final waypoint.
The resulting motion is reported in~\fig{fig:2D_sim} and can be seen in the attached video.
The predicted collision score throughout the trajectory is reported in~\fig{fig:colpred}.
It displays that minor predicted violations of the constraint occur,
which is a result of the input image noise creating sudden changes on the collision map.
This makes the \nmpc solver warm-starting to be far from the optimal solution,
preventing the \ac{RTI} step from properly approximating it.
Those violations are however immediately corrected in subsequent solving steps,
preventing collision scores for the near future to increase.

In particular, we can observe at the end of the corridor that the actual velocity of the \ar is orthogonal to the reference one,
clearly demonstrating a non-greedy behavior.
This illustrates the advantages of utilizing the \nmpc as a local planner,
as its predicting capabilities allow to overcome local minima.

%---------------------------------------------------------------
\subsection{Experiments}\label{subsec:valid:xp}
\subsubsection{System Overview}\label{subsec:valid:xp:system}
\begin{figure}[t]
\centering
    \includegraphics[width=0.99\columnwidth]{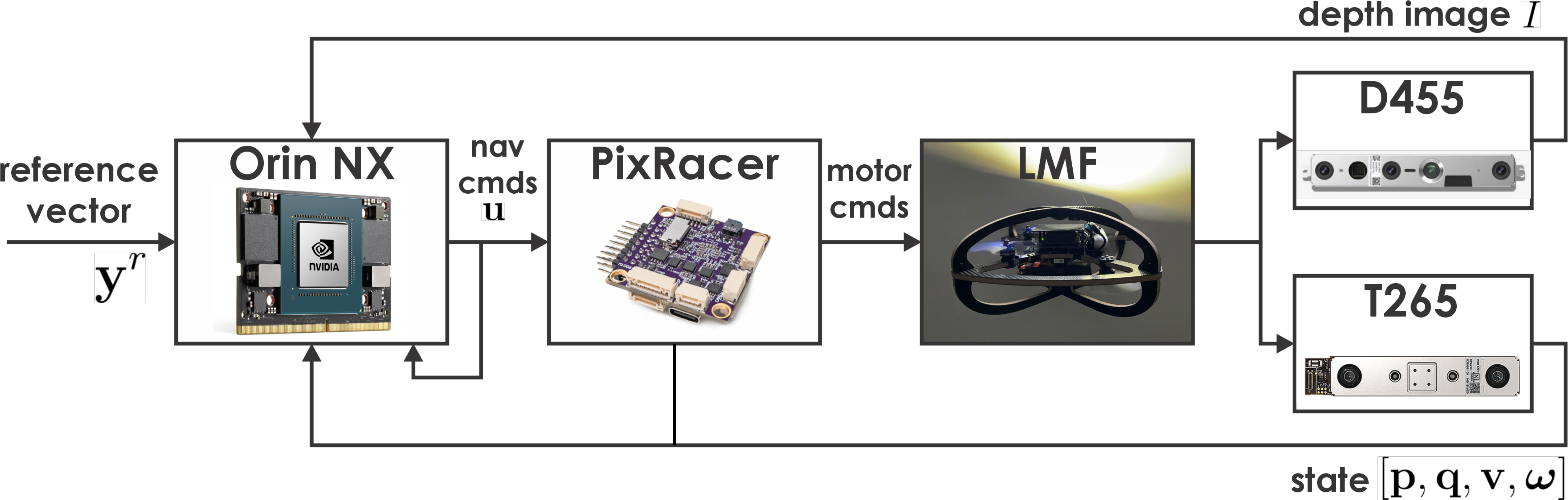}
    \caption{Block diagram of the LMF aerial robot.}
    \label{fig:lmf_hardware}
\end{figure}

To evaluate the method, we utilize the Learning-based Micro Flyer (LMF)~\cite{Nguyen22,Kulkarni23a}, pictured in~\fig{fig:lmf_view}.
The robot has a diameter of $0.43\textrm{m}$ and weights $1.2\textrm{kg}$.
It features a Realsense D455 for depth data at $480\times 270$ pixel resolution and $30$ FPS,
a PixRacer Ardupilot-based autopilot for velocity and yaw-rate control,
and a Realsense T265 fused with the autopilot's IMU for acquiring the robot's odometry state estimate.
The robot integrates an NVIDIA Orin NX onboard in which the proposed method is executed by exploiting its GPU (for the \cnn) and CPU (for the \nmpc).
The system is depicted in Figure~\ref{fig:lmf_hardware}.

\subsubsection{Experimental results}\label{subsec:valid:xp:results}
\begin{figure}[t]
\centering
    \includegraphics[width=0.99\columnwidth]{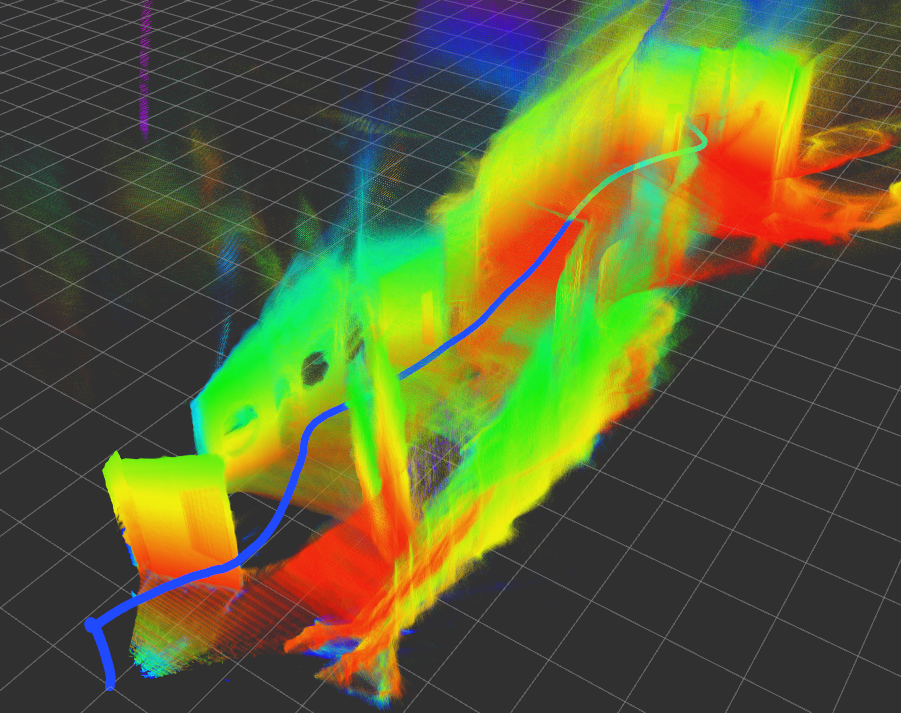}
    \caption{Trajectory (blue line) of LMF avoiding obstacles.
    The goal position is located $15$\unit{m} in front of the starting point.
    The colored dots are the aggregated pointclouds captured by the D455 camera.}
    \label{fig:mission}
\end{figure}

The experiment conducted with LMF consists of reaching a goal position located $15$\unit{m} in front of the starting location.
The reference velocity is $1.5$\unit{m/s}.
Several obstacles are present in the path of the \ar,
such that avoidance maneuvers are required.
The final waypoint is chosen to be behind a wall.
The resulting motion is summarized in~\fig{fig:mission} and can be seen in the attached video.

Table~\ref{tab:cpt} presents the statistics of computation time onboard the Orin NX of our algorithms.
Those data are gather throughout $15$ missions in setups similar to the one showcased in~\fig{fig:mission},
totalizing an aggregated flight time of $10$ minutes.
The core of the method (encoding of depth images and \nmpc solving) runs in less than $10$\unit{ms},
therefore reaching a maximum control frequency of $100$\unit{Hz}.
The \nmpc solving time is consistently around $3$\unit{ms},
with $99\%$ of samples below $4.3$\unit{ms}
although there exist some outliers (with a recorded maximum of $10.3$\unit{ms}).
Despite the depth filling algorithm~\cite{Ku18} being rather slow because working on CPU,
the overall method remains real time and faster than the odometry feedback of the autopilot.

\begin{table}[t]
    \centering
    \footnotesize
    \vspace{1em}
    \resizebox{0.95\columnwidth}{!}{
        \begin{tabular}{|c|c|c|c|}
        \hline
        Process             & depth filling & \cnn  & \nmpc \\
        \hline
        Real images         & 11.85          & 6.85  & 3.09 \\
        \hline
        \end{tabular}
    }
    \caption{Average computation times (in \unit{ms}) of the main parts of the method on the onboard computer.}
    \label{tab:cpt}
\end{table}

%%% CCL %%%%%%%%%%%%%%%%%%%%%%%%%%%%%%%%%%%%%%%%%%%%%%%%%%%%%%%%
\section{Conclusion}\label{sec:ccl}
In this work, we proposed a new paradigm for collision avoidance that merges \nn-based image data processing with \nmpc.
The \nn enables the framework to process in real time depth images that are used by the optimal controller in a cascaded fashion.
This network is divided in two parts.
First, a convolution variational encoder compresses the input image into a latent vector that encompasses collision information.
The second part is a \ac{FC} network which is written as an algebraic function of the \nmpc state,
constraining the predicted position to avoid perceived obstacles.
The proposed method is first quantitatively evaluated on its classification performances,
with emphasis on assessing the performance of the simulation-trained network on real images.
Then the resulting \nmpc controller is evaluated on simulation and experiments,
showcasing avoidance behaviors and high frequency on the onboard computer,
despite utilizing the neural networks.
The one-by-one image processing is however constraining the motion to belong to the current \fov,
preventing exploitation of the full dynamics of the \ar.
Thus, future works include exploring memory-based neural network to encode information from several short-past images,
enabling the \nmpc to plan in a less constrained space.
The injection of real data in the training could also be explored to render the \nn robust to stereo errors,
such as shadowing,
and thus discard the need for a time-consuming and error-prone depth-filling algorithm.
Finally, future works should also provide in-depth performance comparison with other existing sensor-based collision avoidance methods,
as well as map-based path planners.

%%% BIBLIOGRAPHY %%%%%%%%%%%%%%%%%%%%%%%%%%%%%%%%%%%%%%%%%%%%%%%
\bibliographystyle{IEEEtran}
\bibliography{alias, main}
\end{document}

%% file: 0_abbrev.tex
\newcommand\norm[1]{\left\lVert#1\right\rVert}   % norms
\newcommand{\vect}[1]{\mathbf{#1}}  % vectors
\newcommand{\mat}[1]{\vect{#1}}  % matrix
\newcommand{\vecg}[1]{\bm{#1}}  % vector with greek letter
\newcommand{\R}{\mathbb{R}}  % real space
\newcommand{\bmat}[1]{\begin{bmatrix} #1 \end{bmatrix}}  % big matrix
\newcommand{\transp}{^{\top}}  % transpose
\newcommand{\ts}[1]{{}^{\sb{#1}}}  % small topscripts
\newcommand{\us}[1]{{}_{\sb{#1}}}  % small subscripts
\newcommand{\frameV}[1]{\mathcal{F}_{\sb{#1}}}  % reference frame

\newcommand{\interv}[2]{\left[#1,\;#2\right]}  % continuous interval
\newcommand{\intervei}[2]{\left(#1,\;#2\right]}  % cont. int. exclusive left
\newcommand{\intervint}[2]{\left\{#1,\;#2\right\}}  % discrete set

\newcommand{\ie}{i.e.\xspace}
\newcommand{\eg}{e.g.\xspace}

\newcommand{\fig}[1]{Fig.~\ref{#1}\xspace}
\newcommand{\sect}[1]{Sec.~\ref{#1}\xspace}
\newcommand{\tab}[1]{Tab.~\ref{#1}\xspace}
\newcommand{\eqn}[1]{Eq.~\eqref{#1}\xspace}

\newcommand{\tturl}[1]{\tt\url{#1}}

%% file: 0_glossary.tex
\usepackage{acro}

\DeclareAcronym{CBF}{
    short = CBF,
    long = Control Barrier Function
}

\DeclareAcronym{CAD}{
    short = CAD,
    long = computer aided design
}

\DeclareAcronym{RMSE}{
    short = RMSE,
    long = Root Mean Square Error
}

\newcommand{\std}{\ac{std}\xspace}
\DeclareAcronym{std}{
    short = std,
    long = standard deviation
}

\newcommand{\wrt}{\ac{wrt}\xspace}
\DeclareAcronym{wrt}{
    short = w.r.t.,
    long = with respect of
}

\DeclareAcronym{MRS}{
    short = MRS,
    long = Multi-Robots Systems
}

\DeclareAcronym{AIA}{
    short = AIA,
    long = Active Information Acquisition
}

\DeclareAcronym{VIO}{
    short = VIO,
    long = Visual-Inertial Odometry
}
\DeclareAcronym{SLAM}{
    short = SLAM,
    long = Simultaneous Localization And Mapping
}
\DeclareAcronym{VISLAM}{
    short = VISLAM,
    long = Visual-Inertial \ac{SLAM}
}

\newcommand{\nn}{\ac{NN}\xspace}
\newcommand{\nns}{\acp{NN}\xspace}
\DeclareAcronym{NN}{
    short = NN,
    long = Neural Network
}
\DeclareAcronym{FC}{
    short = FC,
    long = Fully Connected
}
\newcommand{\cnn}{\ac{CNN}\xspace}
\DeclareAcronym{CNN}{
    short = CNN,
    long = Convolutional \acl{NN}
}
\DeclareAcronym{YOLO}{
    short = YOLO,
    long = You Only Look Once
}
\DeclareAcronym{BCE}{
    short = BCE,
    long = Binary Cross Entropy
}
\DeclareAcronym{RL}{
    short = RL,
    long = Reinforcement Learning
}
\DeclareAcronym{IL}{
    short = IL,
    long = Imitation Learning
}

\DeclareAcronym{DOF}{
    short = DoF,
    long = Degree of Freedom,
    long-plural-form = Degrees of Freedom
}

\DeclareAcronym{PnP}{
    short = PnP,
    long = Perspective-n-Points
}
\DeclareAcronym{IPPE}{
    short = IPPE,
    long = Infinitesimal Plane-based Pose Estimation
}
\DeclareAcronym{SfM}{
    short = SfM,
    long = Structure from Motion
}

\DeclareAcronym{KF}{
    short = KF,
    long = Kalman Filter
}

\DeclareAcronym{EKF}{
    short = E\acs{KF},
    long = Extended \acl{KF}
}

\DeclareAcronym{ESEKF}{
    short = ES\acs{KF},
    long = Error State \acl{KF}
}
\DeclareAcronym{UKF}{
    short = U\acs{KF},
    long = Unscented \acl{KF}
}

\DeclareAcronym{IKF}{
    short = I\acs{KF},
    long = Intermittent \acl{KF}
}

\newcommand{\fov}{\ac{FOV}\xspace}
\DeclareAcronym{FOV}{
    short = FoV,
    long = Field of View,
    long-plural-form = Fields of View
}

\newcommand{\ar}{\ac{AR}\xspace}
\newcommand{\ars}{\acp{AR}\xspace}
\DeclareAcronym{AR}{
    short = AR,
    long = Aerial Robot
}

\DeclareAcronym{GTMR}{
    short = GTMR,
    long = Generically Tilted Multi-Rotor
}

\DeclareAcronym{UAV}{
    short = UAV,
    long = Uncrewed Aerial Vehicle
}

\DeclareAcronym{UGV}{
    short = UGV,
    long = Uncrewed Ground Vehicle
}

\newcommand{\mpc}{\ac{MPC}\xspace}
\DeclareAcronym{MPC}{
    short = MPC,
    long = Model Predictive Control
}
\newcommand{\nmpc}{\ac{NMPC}\xspace}

\DeclareAcronym{NMPC}{
    short = N-\acs{MPC},
    long = Nonlinear \acl{MPC}
}

\DeclareAcronym{QP}{
    short = QP,
    long = Quadratic programming
}
\DeclareAcronym{SQP}{
    short = SQP,
    long = Sequential Quadratic programming
}

\DeclareAcronym{OCP}{
    short = OCP,
    long = Optimal Control Problem
}

\DeclareAcronym{RTI}{
    short = RTI,
    long = Real-Time Iteration
}
\DeclareAcronym{LQR}{
    short = LQR,
    long = Linear-Quadratic Regulator
}
\newcommand{\nlp}{\ac{NLP}\xspace}
\DeclareAcronym{NLP}{
    short = NLP,
    long = NonLinear Programming
}
\DeclareAcronym{BVP}{
    short = BVP,
    long = Boundary-Value Problem
}
\DeclareAcronym{KKT}{
    short = KKT,
    long = Karush-Kuhn-Tucker
}

\DeclareAcronym{i2c}{
    short = I2C,
    long = Inter-Integrated Circuit
}
\DeclareAcronym{ESC}{
    short = ESC,
    long = Electronic Speed Controller
}
\DeclareAcronym{GPS}{
    short = GPS,
    long = Global Positioning System
}
\DeclareAcronym{RTK}{
    short = RTK-\acs{GPS},
    long = Real-Time Kinematics \acl{GPS}
}
\DeclareAcronym{mocap}{
    short = MoCap,
    long = Motion Capture
}
\DeclareAcronym{IMU}{
    short = IMU,
    long = Inertial Measurement Unit
}

\DeclareAcronym{RGBD}{
    short = RGBD,
    long = RGB + Depth
}
\DeclareAcronym{LIDAR}{
    short = lidar,
    long = LIght Detection And Ranging,
    long-plural = {},
    first-style = short
}

\DeclareAcronym{COM}{
    short = CoM,
    long = Center of Mass
}

\DeclareAcronym{CPU}{
    short = CPU,
    long = Central Processing Unit
}
\DeclareAcronym{GPU}{
    short = GPU,
    long = Graphical Processing Unit
}
\DeclareAcronym{FPGA}{
    short = FPGA,
    long = Field-Programmable Gate Array
}

\DeclareAcronym{VTOL}{
    short = VTOL,
    long = Vertical Take-Off and Landing
}

\DeclareAcronym{DARPA}{
    short = DARPA,
    long = Defense Advanced Research Projects Agency
}
\DeclareAcronym{MBZIRC}{
    short = MBZIRC,
    long = Mohamed bin Zayed International Robotics Challenge
}

\DeclareAcronym{ANR}{
    short = ANR,
    long = National Research Agency
}
\DeclareAcronym{murophen}{
    short = MuRoPhen,
    long = MuRoPhen (Multiple Robots for observing dynamical Phenomena),
    first-style = long
}
\DeclareAcronym{RIS}{
    short = RIS,
    long = Robotics and InteractionS
}
\DeclareAcronym{LAAS}{
    short = LAAS-CNRS,
    long = Laboratory for Analysis and Architecture of Systems of the French CNRS
}

\DeclareAcronym{VS}{
    short = VS,
    long = Visual Servoing
}

\DeclareAcronym{HVS}{
    short = H\acs{VS},
    long = Hybrid \acl{VS}
}
\DeclareAcronym{PBVS}{
    short = PB\acs{VS},
    long = Position-Based \acl{VS}
}
\DeclareAcronym{IBVS}{
    short = IB\acs{VS},
    long = Image-Based \acl{VS}
}

\DeclareAcronym{WO}{
    short = WO,
    long = Wrench Observer
}
\DeclareAcronym{AF}{
    short = AF,
    long = Admittance Filter
}

\DeclareAcronym{RANSAC}{
    short = RANSAC,
    long = RANdom SAmpling Consensus
}

\DeclareAcronym{CW}{
    short = CW,
    long = Clockwise
}

\DeclareAcronym{CCW}{
    short = CCW,
    long = Counter-Clockwise
}

\DeclareAcronym{MAP}{
    short = MAP,
    long = Maximum A Posteriori
}
\DeclareAcronym{ML}{
    short = ML,
    long = Maximum Likelyhood
}

\DeclareAcronym{INDI}{
    short = INDI,
    long = Incremental Nonlinear Dynamic Inversion
}

%% file: imgs/colpred.pdf_tex
%% Creator: Inkscape 1.3 (1:1.3+202307231459+0e150ed6c4), www.inkscape.org
%% Accompanies image file 'colpred.pdf' (pdf, eps, ps)
\begingroup%
  \makeatletter%
  \providecommand\color[2][]{%
    \errmessage{(Inkscape) Color is used for the text in Inkscape, but the package 'color.sty' is not loaded}%
    \renewcommand\color[2][]{}%
  }%
  \providecommand\transparent[1]{%
    \errmessage{(Inkscape) Transparency is used (non-zero) for the text in Inkscape, but the package 'transparent.sty' is not loaded}%
    \renewcommand\transparent[1]{}%
  }%
  \providecommand\rotatebox[2]{#2}%
  \newcommand*\fsize{\dimexpr\f@size pt\relax}%
  \newcommand*\lineheight[1]{\fontsize{\fsize}{#1\fsize}\selectfont}%
  \ifx\svgwidth\undefined%
    \setlength{\unitlength}{417.6058167bp}%
    \ifx\svgscale\undefined%
      \relax%
    \else%
      \setlength{\unitlength}{\unitlength * \real{\svgscale}}%
    \fi%
  \else%
    \setlength{\unitlength}{\svgwidth}%
  \fi%
  \global\let\svgwidth\undefined%
  \global\let\svgscale\undefined%
  \makeatother%
  \begin{picture}(1,0.40036337)%
    \lineheight{1}%
    \setlength\tabcolsep{0pt}%
    \put(0,0){\includegraphics[width=\unitlength,page=1]{colpred.pdf}}%
    \put(0,0){\includegraphics[width=\unitlength,page=2]{colpred.pdf}}%
    \put(0,0){\includegraphics[width=\unitlength,page=3]{colpred.pdf}}%
    \put(0.03518871,0.33696809){\color[rgb]{0,0,0}\makebox(0,0)[lt]{\lineheight{1.25}\smash{\begin{tabular}[t]{l}$I$\end{tabular}}}}%
    \put(0.19648208,0.25570548){\color[rgb]{0,0,0}\makebox(0,0)[lt]{\lineheight{1.25}\smash{\begin{tabular}[t]{l}ResNet8\end{tabular}}}}%
    \put(0.46129262,0.32693418){\color[rgb]{0,0,0}\makebox(0,0)[lt]{\lineheight{1.25}\smash{\begin{tabular}[t]{l}$\mu$\end{tabular}}}}%
    \put(0.45978149,0.20527687){\color[rgb]{0,0,0}\makebox(0,0)[lt]{\lineheight{1.25}\smash{\begin{tabular}[t]{l}$\sigma$\end{tabular}}}}%
    \put(0.6080082,0.26710325){\color[rgb]{0,0,0}\makebox(0,0)[lt]{\lineheight{1.25}\smash{\begin{tabular}[t]{l}$\vect{z}$\end{tabular}}}}%
    \put(0.36000039,0.06007617){\color[rgb]{0,0,0}\makebox(0,0)[lt]{\lineheight{1.25}\smash{\begin{tabular}[t]{l}$\vect{a}$\end{tabular}}}}%
    \put(0.48357083,0.05990721){\color[rgb]{0,0,0}\makebox(0,0)[lt]{\lineheight{1.25}\smash{\begin{tabular}[t]{l}FC\end{tabular}}}}%
    \put(0.77180093,0.18934209){\color[rgb]{0,0,0}\makebox(0,0)[lt]{\lineheight{1.25}\smash{\begin{tabular}[t]{l}FC\end{tabular}}}}%
    \put(0.89219338,0.19167739){\color[rgb]{0,0,0}\makebox(0,0)[lt]{\lineheight{1.25}\smash{\begin{tabular}[t]{l}$\hat{c}$\end{tabular}}}}%
    \put(0.51157261,0.26516507){\color[rgb]{0,0,0}\makebox(0,0)[lt]{\lineheight{1.25}\smash{\begin{tabular}[t]{l}$\epsilon$\end{tabular}}}}%
    \put(0.1974705,0.20822901){\color[rgb]{0,0,0}\makebox(0,0)[lt]{\lineheight{1.25}\smash{\begin{tabular}[t]{l}\tiny ReLU\end{tabular}}}}%
    \put(0.48428273,0.03555674){\color[rgb]{0,0,0}\makebox(0,0)[lt]{\lineheight{1.25}\smash{\begin{tabular}[t]{l}\tiny tanh\end{tabular}}}}%
    \put(0.77298136,0.16283097){\color[rgb]{0,0,0}\makebox(0,0)[lt]{\lineheight{1.25}\smash{\begin{tabular}[t]{l}\tiny tanh \end{tabular}}}}%
    \put(0.19819252,0.187214){\color[rgb]{0,0,0}\makebox(0,0)[lt]{\lineheight{1.25}\smash{\begin{tabular}[t]{l}\tiny batchnorm\end{tabular}}}}%
  \end{picture}%
\endgroup%

%% file: main.bbl
\begin{thebibliography}{10}
\providecommand{\url}[1]{#1}
\csname url@samestyle\endcsname
\providecommand{\newblock}{\relax}
\providecommand{\bibinfo}[2]{#2}
\providecommand{\BIBentrySTDinterwordspacing}{\spaceskip=0pt\relax}
\providecommand{\BIBentryALTinterwordstretchfactor}{4}
\providecommand{\BIBentryALTinterwordspacing}{\spaceskip=\fontdimen2\font plus
\BIBentryALTinterwordstretchfactor\fontdimen3\font minus
  \fontdimen4\font\relax}
\providecommand{\BIBforeignlanguage}[2]{{%
\expandafter\ifx\csname l@#1\endcsname\relax
\typeout{** WARNING: IEEEtran.bst: No hyphenation pattern has been}%
\typeout{** loaded for the language `#1'. Using the pattern for}%
\typeout{** the default language instead.}%
\else
\language=\csname l@#1\endcsname
\fi
#2}}
\providecommand{\BIBdecl}{\relax}
\BIBdecl

\bibitem{Tian20}
Y.~Tian, K.~Liu, K.~Ok, L.~Tran, D.~Allen, N.~Roy, and J.~P. How, ``Search and
  rescue under the forest canopy using multiple {UAVs},'' \emph{The Int.
  Journal of Robotics Research}, vol.~39, no. 10-11, pp. 1201--1221, 2020.

\bibitem{Dang20}
T.~Dang, M.~Tranzatto, S.~Khattak, F.~Mascarich, K.~Alexis, and M.~Hutter,
  ``Graph-based subterranean exploration path planning using aerial and legged
  robots,'' \emph{Journal of Field Robotics}, vol.~37, no.~8, pp. 1363--1388,
  2020.

\bibitem{Petracek20}
P.~Petracek, V.~Kratky, and M.~Saska, ``Dronument: System for reliable
  deployment of micro aerial vehicles in dark areas of large historical
  monuments,'' \emph{IEEE Robotics and Automation Letters}, vol.~5, pp.
  2078--2085, 2020.

\bibitem{Zhang19}
P.~Zhang, Y.~Zhong, and X.~Li, ``{SlimYOLOv3}: Narrower, faster and better for
  real-time {UAV} applications,'' in \emph{2020 IEEE/CVF Int. Conf. on Computer
  Vision}, 2019.

\bibitem{Akbari21}
Y.~Akbari, N.~Almaadeed, S.~Al-maadeed, and O.~Elharrouss, ``Applications,
  databases and open computer vision research from drone videos and images: a
  survey,'' \emph{Artificial Intelligence Review}, vol.~54, no.~5, pp.
  3887--3938, 2021.

\bibitem{Khattak20}
S.~Khattak, H.~Nguyen, F.~Mascarich, T.~Dang, and K.~Alexis, ``Complementary
  multi--modal sensor fusion for resilient robot pose estimation in
  subterranean environments,'' in \emph{2020 Int. Conf. on Unmanned Aircraft
  Systems}, 2020, pp. 1024--1029.

\bibitem{Lopez17}
B.~T. Lopez and J.~P. How, ``Aggressive 3-d collision avoidance for high-speed
  navigation.'' in \emph{2017 IEEE Int. Conf. on Robotics and Automation},
  2017, pp. 5759--5765.

\bibitem{Loquercio18}
A.~Loquercio, A.~I. Maqueda, C.~R. Del-Blanco, and D.~Scaramuzza, ``Dronet:
  Learning to fly by driving,'' \emph{IEEE Robotics and Automation Letters},
  vol.~3, no.~2, pp. 1088--1095, 2018.

\bibitem{Loquercio21}
A.~Loquercio, E.~Kaufmann, R.~Ranftl, M.~M{\"u}ller, V.~Koltun, and
  D.~Scaramuzza, ``Learning high-speed flight in the wild,'' \emph{Science
  Robotics}, vol.~6, no.~59, 2021.

\bibitem{Tolani21}
V.~Tolani, S.~Bansal, A.~Faust, and C.~Tomlin, ``Visual navigation among humans
  with optimal control as a supervisor,'' \emph{IEEE Robotics and Automation
  Letters}, vol.~6, no.~2, pp. 2288--2295, 2021.

\bibitem{Kahn21}
G.~Kahn, P.~Abbeel, and S.~Levine, ``{BADGR}: An autonomous self-supervised
  learning-based navigation system,'' \emph{IEEE Robotics and Automation
  Letters}, vol.~6, no.~2, pp. 1312--1319, 2021.

\bibitem{Ugurlu22}
H.~I. Ugurlu, X.~H. Pham, and E.~Kayacan, ``Sim-to-real deep reinforcement
  learning for safe end-to-end planning of aerial robots,'' \emph{Robotics},
  vol.~11, no.~5, p. 109, 2022.

\bibitem{Hoeller21}
D.~Hoeller, L.~Wellhausen, F.~Farshidian, and M.~Hutter, ``Learning a state
  representation and navigation in cluttered and dynamic environments,''
  \emph{IEEE Robotics and Automation Letters}, vol.~6, no.~3, pp. 5081--5088,
  2021.

\bibitem{Nguyen22}
H.~Nguyen, S.~H. Fyhn, P.~De~Petris, and K.~Alexis, ``Motion primitives-based
  navigation planning using deep collision prediction,'' in \emph{2022 IEEE
  Int. Conf. on Robotics and Automation}, 2022, pp. 9660--9667.

\bibitem{Kaufmann23}
E.~Kaufmann, L.~Bauersfeld, A.~Loquercio, M.~M{\"u}ller, V.~Koltun, and
  D.~Scaramuzza, ``Champion-level drone racing using deep reinforcement
  learning,'' \emph{Nature}, vol. 620, no. 7976, pp. 982--987, 2023.

\bibitem{Tan23}
\BIBentryALTinterwordspacing
D.~C. Tan, F.~Acero, R.~McCarthy, D.~Kanoulas, and Z.~A. Li, ``Value functions
  are control barrier functions: Verification of safe policies using control
  theory,'' in \emph{2nd Workshop on Formal Verification and Machine Learning},
  2023. [Online]. Available: \url{https://arxiv.org/abs/2306.04026}
\BIBentrySTDinterwordspacing

\bibitem{Dawson22}
C.~Dawson, B.~Lowenkamp, D.~Goff, and C.~Fan, ``Learning safe, generalizable
  perception-based hybrid control with certificates,'' \emph{IEEE Robotics and
  Automation Letters}, vol.~7, no.~2, pp. 1904--1911, 2022.

\bibitem{Williams17}
G.~Williams, N.~Wagener, B.~Goldfain, P.~Drews, J.~M. Rehg, B.~Boots, and E.~A.
  Theodorou, ``Information theoretic {MPC} for model-based reinforcement
  learning,'' 2017, pp. 1714--1721.

\bibitem{Chee22}
K.~Y. Chee, T.~Z. Jiahao, and M.~A. Hsieh, ``{KNODE-MPC}: A knowledge-based
  data-driven predictive control framework for aerial robots,'' \emph{IEEE
  Robotics and Automation Letters}, vol.~7, no.~2, pp. 2819--2826, 2022.

\bibitem{Salzmann23}
T.~Salzmann, E.~Kaufmann, J.~Arrizabalaga, M.~Pavone, D.~Scaramuzza, and
  M.~Ryll, ``Real-time neural {MPC}: Deep learning model predictive control for
  quadrotors and agile robotic platforms,'' \emph{IEEE Robotics and Automation
  Letters}, vol.~8, no.~4, pp. 2397--2404, 2023.

\bibitem{Mescheder19}
L.~Mescheder, M.~Oechsle, M.~Niemeyer, S.~Nowozin, and A.~Geiger, ``Occupancy
  networks: Learning 3d reconstruction in function space,'' in \emph{2019
  IEEE/CVF Conf. on Computer Vision and Pattern Recognition}, 2019, pp.
  4460--4470.

\bibitem{Tancik20}
M.~Tancik, P.~Srinivasan, B.~Mildenhall, S.~Fridovich-Keil, N.~Raghavan,
  U.~Singhal, R.~Ramamoorthi, J.~Barron, and R.~Ng, ``Fourier features let
  networks learn high frequency functions in low dimensional domains,'' pp.
  7537--7547, 2020.

\bibitem{Higgins17}
I.~Higgins, L.~Matthey, A.~Pal, C.~Burgess, X.~Glorot, M.~Botvinick,
  S.~Mohamed, and A.~Lerchner, ``$\beta$-{VAE}: Learning basic visual concepts
  with a constrained variational framework,'' in \emph{2017 Int. Conf. on
  Learning Representations}, 2017.

\bibitem{Kulkarni23b}
\BIBentryALTinterwordspacing
M.~Kulkarni, T.~J. Forgaard, and K.~Alexis, ``Aerial gym--isaac gym simulator
  for aerial robots,'' in \emph{"The Role of Robotics Simulators for Unmanned
  Aerial Vehicles" Workshop at 2023 IEEE Int. Conf. on Robotics and
  Automation}, 2023. [Online]. Available:
  \url{https://arxiv.org/abs/2305.16510}
\BIBentrySTDinterwordspacing

\bibitem{Kulkarni23c}
M.~Kulkarni and K.~Alexis, ``Task-driven compression for collision encoding
  based on depth images,'' in \emph{2023 Int. Symp. on Visual Computing}, 2023.

\bibitem{Kulkarni23a}
M.~Kulkarni, H.~Nguyen, and K.~Alexis, ``Semantically-enhanced deep collision
  prediction for autonomous navigation using aerial robots,'' in \emph{2023
  IEEE/RSJ Int. Conf. on Intelligent Robots and Systems}, 2023.

\bibitem{Ku18}
J.~Ku, A.~Harakeh, and S.~L. Waslander, ``In defense of classical image
  processing: Fast depth completion on the {CPU},'' in \emph{2018 15th Conf. on
  Computer and Robot Vision}, 2018.

\bibitem{AcadosLib}
R.~Verschueren, G.~Frison, D.~Kouzoupis, J.~Frey, N.~van Duijkeren, A.~Zanelli,
  B.~Novoselnik, T.~Albin, R.~Quirynen, and M.~Diehl, ``acados -- a modular
  open-source framework for fast embedded optimal control,'' \emph{Mathematical
  Programming Computation}, vol.~14, p. 147–183, 2021.

\bibitem{CasADiLib}
J.~A. Andersson, J.~Gillis, G.~Horn, J.~B. Rawlings, and M.~Diehl, ``{CasADi}:
  a software framework for nonlinear optimization and optimal control,''
  \emph{Mathematical Programming Computation}, vol.~11, no.~1, pp. 1--36, 2019.

\bibitem{Mallet10}
A.~Mallet, C.~Pasteur, M.~Herrb, S.~Lemaignan, and F.~Ingrand, ``{GenoM3}:
  Building middleware-independent robotic components,'' in \emph{2010 IEEE Int.
  Conf. on Robotics and Automation}, 2010, pp. 4627--4632.

\end{thebibliography}
